\documentclass[letterpaper, 10 pt, conference]{ieeeconf}  

\IEEEoverridecommandlockouts                              

\usepackage[T1]{fontenc}

\usepackage{amsmath,amssymb,mathtools,bm}
\usepackage{newtxtext}
\usepackage{newtxmath}

\usepackage{graphicx}
\usepackage{booktabs}
\usepackage{multirow}
\usepackage{array}
\usepackage{subcaption}

\usepackage[table]{xcolor}
\definecolor{darkred}{rgb}{0.39,0.04,0.45}
\definecolor{cvprblue}{rgb}{0.21,0.49,0.74}

\let\labelindent\relax
\usepackage{enumitem}
\setlist{
    nosep,
    leftmargin=*,
    itemsep=0.2ex,
    topsep=0.2ex
}

\usepackage[ruled,vlined]{algorithm2e}

\usepackage{pifont}
\usepackage{afterpage}

\usepackage[
    pagebackref,
    breaklinks,
    colorlinks,
    allcolors=cvprblue
]{hyperref}

\title{ChunkFlow: Towards Continuity-Consistent Chunked Policy Learning}

\author{Zhao Yang$^{1,2}$, Yinan Shi$^{2}$, Mingyuan Yao$^{2}$, Wenyao Xue$^{2}$, Yawei Jueluo$^{2}$ and Longjun Liu$^{1,*}$%
\thanks{$^{1}$State Key Laboratory of Human-Machine Hybrid Augmented Intelligence, Institute of Artificial Intelligence and Robotics, Xi'an Jiaotong University, China.}%
\thanks{$^{2}$Jiangsu Cytoderm Intelligent Technology Co., Ltd., China.}%
\thanks{$^{*}$Longjun Liu is the corresponding author.}%
}

\begin{document}

\maketitle
\thispagestyle{empty}
\pagestyle{empty}

\begin{abstract}
Vision–language action (VLA) models increasingly adopt chunked action heads to satisfy real-time constraints; however, this introduces \textit{boundary jitter}: overlapping regions between consecutive chunks often yield inconsistent predictions, degrading temporal coherence and the task success rate. Existing methods, such as inference-time blending, merely reweight mismatched proposals without correcting underlying errors, leading to residual accumulation under biased or noisy histories. We propose {ChunkFlow}, a seam-aware training-and-execution framework for chunked policies that aligns chunk structure with boundary execution. It partitions each chunk into frozen, editable, and future zones, applies deterministic overlap blending at execution, and trains raw predictions with seam and first- and second-order continuity losses. History corruption and scheduled sampling improve robustness to executed-history errors, while an AWAC fine-tuning stage adapts the policy without removing these structural regularizers. Under mild smoothness assumptions, pre-blending seam discrepancies provably decay with increasing overlap. Experiments on CALVIN, LIBERO, and real robots show an improved success-stability trade-off with low-latency inference. Project page: \url{https://cytoderm-ai.github.io/chunkflow}.
\end{abstract}

\section{Introduction}

Recent advances in robotic manipulation increasingly leverage vision–language action (VLA) models to translate open-ended instructions into executable behaviors. To enable real-time control, modern VLA architectures typically employ \textit{action-chunking} heads, which emit short action sequences (chunks) per decision step~\cite{jiang2023vima, cheang2024gr2generativevideolanguageactionmodel, kim2024openvla, ghosh2024octo, huang2024robotfp}. This design amortizes inference costs and improves deployment efficiency, but it introduces a critical failure mode: \textit{temporal discontinuities at chunk boundaries}. Since each chunk is predicted under slightly shifted observations and histories, overlaps between chunks often yield conflicting actions—causing jitter, degraded coherence, and task failure. This raises a fundamental question: how can we retain task effectiveness while ensuring smooth, consistent transitions across chunked predictions?

Most existing mitigations handle boundary artifacts heuristically at inference time or overlook execution semantics during training. RTC~\cite{black2025real} applies online interpolation to smooth seams, but such \textbf{inference-only blending} reweights misaligned chunk predictions without correcting upstream errors, causing residuals to accumulate under noisy or biased histories. 
Moreover, the absence of \textbf{training-time seam supervision} yields weak boundary gradients, preventing policies from internalizing blendable structure. 
Mainstream VLA pipelines—e.g., OpenVLA~\cite{kim2024openvla}, GR2~\cite{cheang2024gr2generativevideolanguageactionmodel}, TraceVLA~\cite{zheng2024tracevla}—apply uniform chunk-wise losses and ignore execution-indexed semantics (frozen, editable, future), leaving boundary inconsistencies unresolved. 
Other approaches explore chunked control via flow/diffusion models~\cite{chi2024universal,black2024pi}, skill abstraction~\cite{lee2024behavior,mete2024questselfsupervisedskillabstractions}, or behavior priors~\cite{liu2024rdt}; while effective offline, these lack \textbf{executed-history feedback} and seam-aware learning, limiting deployment-time alignment.

\begin{figure}[t]
\centering

\includegraphics[width=0.24\linewidth, height=0.2\linewidth]{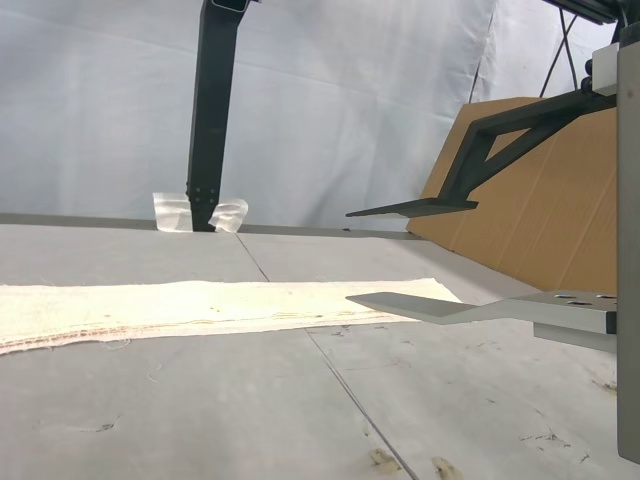}
\includegraphics[width=0.24\linewidth, height=0.2\linewidth]{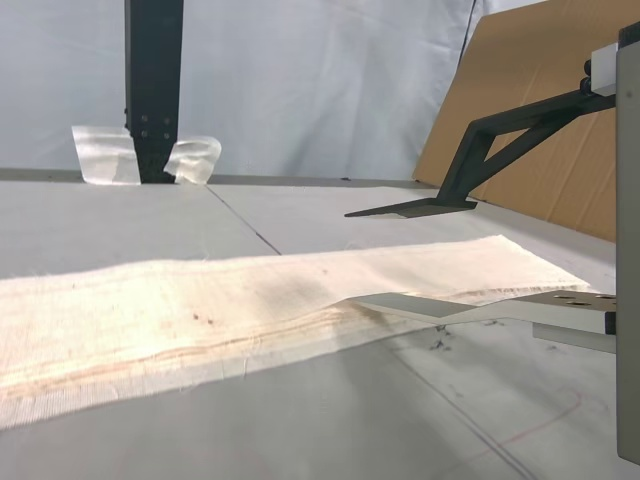}
\includegraphics[width=0.24\linewidth, height=0.2\linewidth]{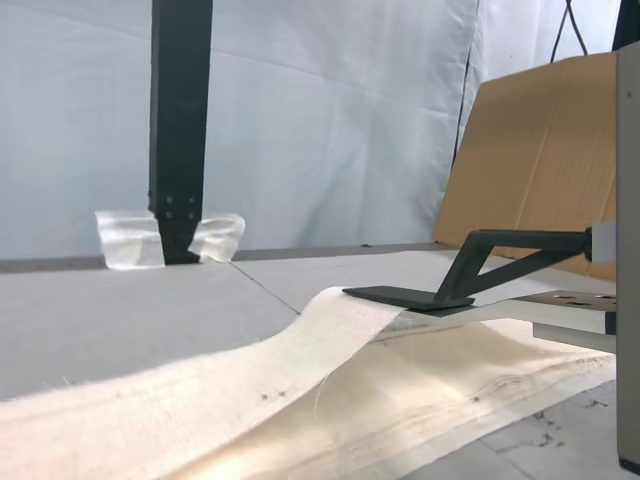}
\includegraphics[width=0.24\linewidth, height=0.2\linewidth]{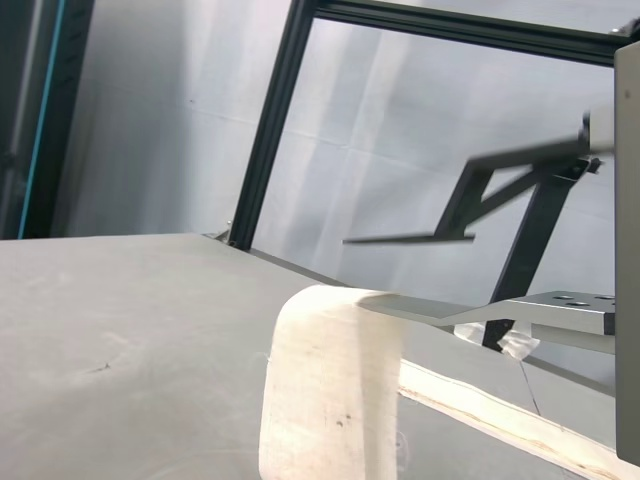}

\includegraphics[width=1.0\linewidth, height=0.22\linewidth]{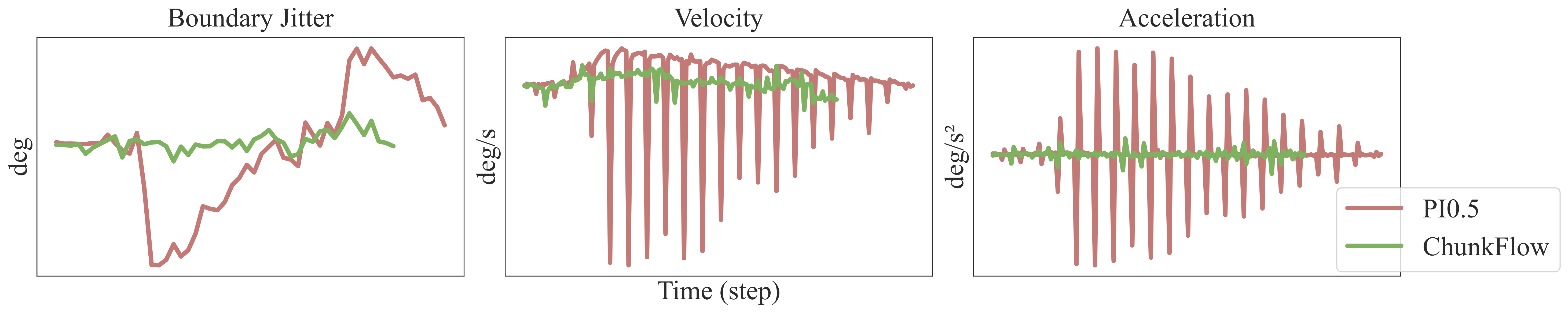}

\caption{\footnotesize
\textbf{Real-world rollout on \textit{Strip-cloth grasping}.}
We deploy \textsc{ChunkFlow} on a manipulation task where the robot extracts a strip from a cloth stack and places it at a target location. \textbf{Top:} keyframes. \textbf{Bottom:} EE roll-angle traces over the first 200 steps, labeled as \textit{jitter} $\Delta a$, \textit{velocity} $\Delta^2 a$, and \textit{acceleration} $\Delta^3 a$ (finite differences of the EE roll angle). Compared to PI0.5~\cite{black2025pi05}, \textsc{ChunkFlow} reduces boundary-induced spikes and high-frequency artifacts across all three signals (consistent with lower MSD-$\Delta a$, MSD-$\Delta^2 a$, and MSD-$\Delta^3 a$), supporting smoother execution.}
\label{fig:teaser}
\end{figure}

In this work, we present \textsc{ChunkFlow}, a seam-aware training-and-execution framework for chunked policies with three integrated components.
\textcolor{darkred}{\ding{182}} \textbf{Structure-aligned overlap blending.}
Each chunk is divided into \emph{frozen}, \emph{editable}, and \emph{future} zones, and adjacent chunks are deterministically reconciled without an extra policy forward pass. Unlike inference-only smoothing~\cite{black2025real}, the overlap structure is exposed during training through pre-blending seam supervision.
\textcolor{darkred}{\ding{183}} \textbf{Continuity-regularized policy optimization.}
First- and second-order penalties and a raw seam loss encourage temporally consistent predictions, while history corruption and scheduled sampling reduce exposure bias.
\textcolor{darkred}{\ding{184}} \textbf{Structure-preserving advantage-weighted fine-tuning.}
AWAC~\cite{nair2020awac} with expectile critics~\cite{kostrikov2021iql} adapts the policy from post-blended histories while retaining seam and continuity regularization. Under mild smoothness assumptions, seam discrepancy decreases with overlap. Thus, the contribution lies in seam-aware training and execution for chunked policies, rather than VLA backbone design.

We evaluate \textsc{ChunkFlow} on CALVIN~\cite{mees2022calvin}, LIBERO~\cite{zhu2023libero}, and two real-robot tasks on a self-developed arm~\cite{cytoderm2025}. Across these settings, \textsc{ChunkFlow} improves the success--smoothness trade-off while retaining low-latency execution, reaching an average sequence length of 4.30 on CALVIN, 93.4\% success rate on LIBERO, 4.43\,ms amortized reasoning latency, and 9/10 successes on the real-robot tasks.

\begin{figure*}[t]
  \centering
  \includegraphics[width=\linewidth]{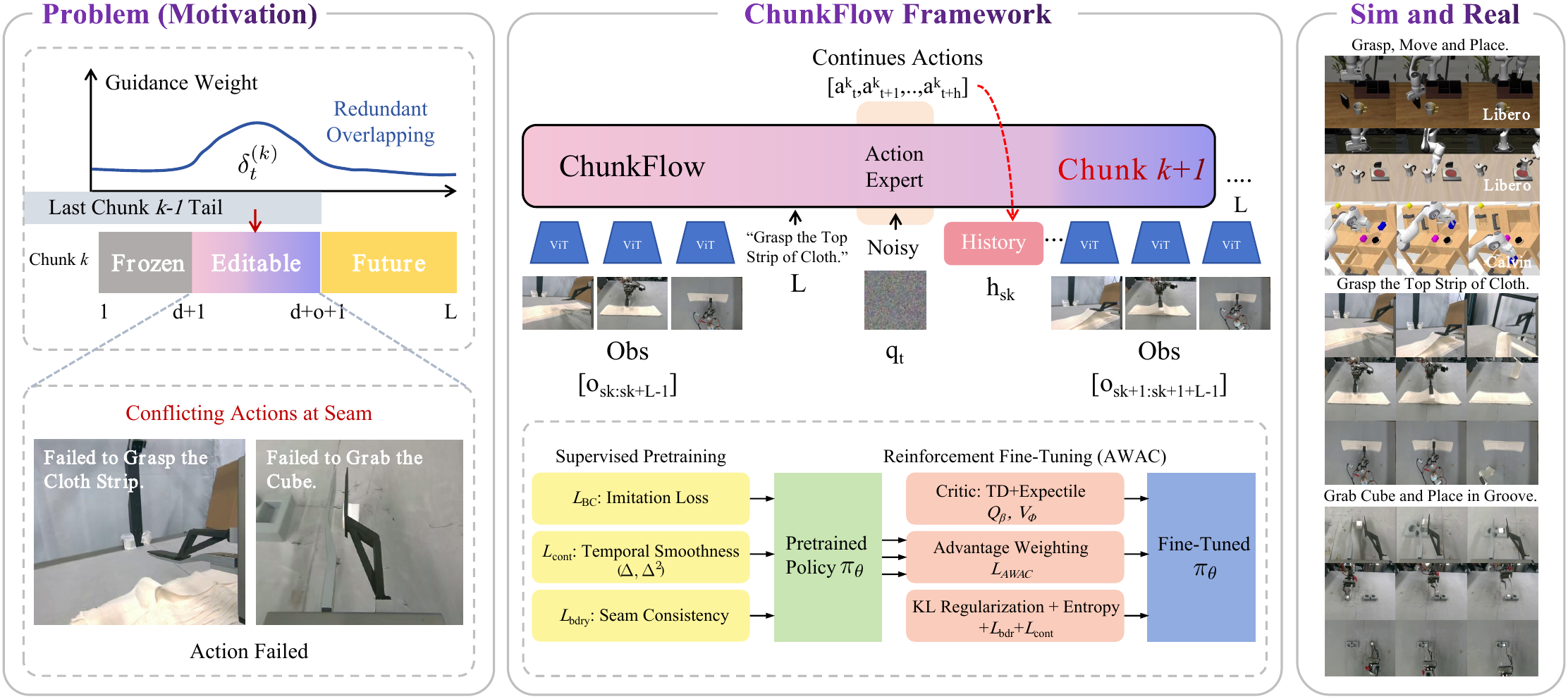}
\caption{
  \textbf{Overview of ChunkFlow.}
  The policy generates temporally overlapping action chunks, which are deterministically blended at boundaries to ensure seam continuity.
  Training integrates supervised regularization and reinforcement fine-tuning for stable long-horizon execution.
  Each predicted chunk is partitioned into three segments: \textcolor{gray}{\emph{frozen}} (gray, length $d$), \textcolor{violet}{\emph{editable seam}} (purple-to-pink gradient, length $O$), and \textcolor{orange}{\emph{future}} (yellow, length $s$), corresponding to index ranges $[1{:}d]$, $[d{+}1{:}d{+}O]$, and $[d{+}O{+}1{:}L]$, respectively.
}
  \label{fig:chunkflow_overview}
  \vspace{-0.4cm}
\end{figure*}
\section{Related Work}
\subsection{Action Chunking and VLA}
Chunked action generation is widely adopted in visuomotor learning for improving long-horizon consistency~\cite{chi2024universal,pertsch2025fast,zhao2024aloha}, often using generative backbones—diffusion and flow models~\cite{black2024pi,braun2024riemannian}, vector quantization~\cite{lee2024behavior,belkhale2024minivla}, or token-based priors~\cite{liu2024rdt}—to model long-range dependencies, including extensions via world models and temporal abstractions~\cite{shao2024genie,zhou2025dreamwalker,mendonca2024adapose}. Despite success offline, most methods overlook \emph{seam inconsistencies} in overlapping regions, causing jitter and instability during deployment. Similarly, large-scale VLA models like OpenVLA, GR2, TraceVLA, and Seer~\cite{kim2024openvla,cheang2024gr2generativevideolanguageactionmodel,zheng2024tracevla} scale instruction-conditioned control with vision–language pretraining~\cite{khazatsky2024droid,walke2023bridgedata,huang2024robotfp} but lack explicit handling of execution-time continuity. We address these gaps via a \emph{seam-aware chunked execution layer} beneath the VLA encoder, aligning chunk boundaries through \emph{deterministic overlap blending} and a \emph{boundary consistency loss} without modifying the multimodal front-end.

\subsection{Temporal Alignment and Smoothness for Chunked Policy Execution}
\label{sec:related_temporal_rl}
Temporal consistency is critical for stable long-horizon control. Prior work enforces intra-chunk smoothness using total-variation, curvature, or jerk penalties~\cite{pertsch2025fast,chi2024universal,braun2024riemannian,zhao2024aloha}, and generative planners~\cite{black2024pi,cheang2024gr2generativevideolanguageactionmodel,cheng2024navila,jang2024tempodp} implicitly regularize motion through extended temporal context. Other efforts explicitly introduce smooth policy updates or temporally consistent rollouts in RL~\cite{wang2024smoothrl,agarwal2024consistent}, while hierarchical behavior regularizers~\cite{yuan2023hbr} further reduce high-frequency artifacts. However, these approaches largely overlook \emph{seam mismatches}—conflicts in overlapping regions—that disrupt rollout stitching in chunked controllers. Imitation-and-RL pipelines~\cite{chi2024universal,zhao2024aloha,lee2024behavior} adopt advantage-weighted updates, expectile critics~\cite{black2024pi,pertsch2025fast}, or latent KL constraints~\cite{zheng2024tracevla,liu2024rdt,kim2024openvla}; yet they still ignore overlap semantics. In contrast, we embed \emph{seam-aware execution} as a structural prior, aligning BC and RL updates with deterministic overlap blending and boundary-aware continuity losses to ensure stable long-horizon behavior.

\section{Method}
\label{sec:method}



\subsection{Problem Setup}
\label{sec:problem_setup}
We study instruction-conditioned control in continuous action spaces. At each decision index $k$, a chunked policy outputs an $L$-step action sequence:
\begin{equation}
a^{(k)}_{1:L} \sim \pi_\theta\!\left(\cdot~\middle|~o_{s_k:s_k+L-1},~h_{s_k},~l\right),
\end{equation}
where $a^{(k)}_t \in \mathbb{R}^d$, $h_{s_k} = (a^{(k)}_{s_k-p}, \dots, a^{(k)}_{s_k-1})$ is a $p$-step action history (zero-padded at the start of the episode), and $s_k = 1 + (k{-}1)S$ with a stride of $S = L{-}O$. Overlapping chunk boundaries ($O{>}0$) allow for smoother transitions but can also introduce redundant or inconsistent predictions in shared steps.

We define the seam discrepancy in the overlapping region as
\begin{equation}
\delta^{(k)}_t = a^{(k)}_t - a^{(k{-}1)}_{t+L-O}, \qquad t = 1, \dots, O.
\end{equation}
The objective is to learn a chunked policy $\pi_\theta$ that imitates expert behavior under instruction $l$ while minimizing $\delta^{(k)}_t$, yielding coherent and smooth action streams for long-horizon execution.

\subsection{Continuity-Consistent Policy Execution via Chunked Action Sequences}
\label{sec:framework_overview}
\textsc{ChunkFlow} is a seam-aware policy framework that generates temporally extended action chunks with enforced continuity across overlapping regions. At each index $k$, the policy $\pi_\theta$ predicts a chunk of $L$ actions $a^{(k)}_{1:L}$, conditioned on local observations $o_{s_k:s_k+L-1}$, a short action history $h_{s_k}$, and high-level instruction $l$. Consecutive chunks overlap for $O$ steps, allowing deterministic blending at the seam to align the head of chunk $k$ with the tail of chunk $k{-}1$ (Fig.~\ref{fig:chunkflow_overview}). To promote long-horizon stability, training includes first- and second-order smoothness regularization and history perturbation for robustness. The policy is initialized via behavior cloning under these constraints and is further refined with advantage-weighted updates while preserving seam structure. During inference, each chunk commits its first $S{=}L{-}O$ steps, blending the overlap and executing the rest directly, enabling low-latency rollout with smooth transitions and adaptive execution.

\subsection{Deterministic Overlap Blending for Seamless Chunk Transitions}
\label{sec:overlap_blending}
Chunked inference improves planning granularity but often introduces discontinuities in overlapping regions between consecutive chunks, resulting in trajectory jitter that undermines real-time stability. We propose a parameter-free \emph{overlap blending} mechanism to enforce smooth transitions without introducing latency. Let $a^{(k-1)}_t$ and $a^{(k)}_t$ denote the predicted actions from chunks $k{-}1$ and $k$ in an overlapping region of length $O$. The executed action at time $s_k{+}t{-}1$ is:
\begin{equation}
\tilde{a}_{s_k + t - 1} = w_t\,a^{(k)}_t + (1 - w_t)\,a^{(k-1)}_{t + L - O}, \quad t = 1, \dots, O,
\label{eq:blend}
\end{equation}
where the interpolation weight $w_t \in [0,1]$ is as follows:
\begin{equation}
w_t = \begin{cases}
0, & O = 1,\\[4pt]
\frac{t - 1}{O - 1}, & O > 1
\end{cases}
\quad\text{(use raw $a^{(k)}_{1:L}$ when $O = 0$)}.
\label{eq:overlap_weight}
\end{equation}
This linearly shifts control from the prior chunk to the current one, requiring only $\mathcal{O}(d)$ flops per overlap step.

To enforce consistency during training, we introduce a boundary loss:
\begin{equation}
L_{\text{bdry}} = \lambda_B \sum_{t=1}^{O} \left\| a^{(k)}_t - a^{(k-1)}_{t + L - O} \right\|_2^2,
\end{equation}
which penalizes seam discrepancies $\delta^{(k)}_t = a^{(k)}_t - a^{(k-1)}_{t + L - O}$ (cf. Sec.~\ref{sec:problem_setup}). This regularization improves pre-blending agreement and enhances rollout continuity. As blending is deterministic and latency-free, it remains suitable for real-time execution and mitigates residual boundary artifacts at test time.


\noindent\textbf{Theoretical Justification.}  
Under local Lipschitz continuity of $\pi_\theta$ and motion-proportional input drift, the expected seam error satisfies:
\begin{equation}
\mathbb{E}\!\left[ \big\| \delta^{(k)}_t \big\|_2^2 \right] = \mathcal{O}((L - O)^2),
\end{equation}
i.e., increasing $O$ quadratically suppresses boundary mismatch, consistent with empirical results in Table~\ref{tab:ablation_hyper}.

\subsection{History-Conditioned Policy with Continuity Regularization}
\label{sec:history_policy}
While overlap blending smooths transitions at chunk boundaries, it does not reduce prediction errors—only interpolates conflicting outputs. When the policy depends on recent action history for temporal context, residual errors can accumulate across chunks, especially under noisy or biased inputs. To mitigate this, we adopt a history-conditioned formulation where the policy receives the current observation $o_t$, an ordered $p$-step history $h_{s_k} = (a^{(k)}_{s_k-p}, \dots, a^{(k)}_{s_k-1})$, and a language instruction~$l$:
\begin{equation}
a^{(k)}_t = \pi_\theta(o_t, h_{s_k}, l).
\end{equation}
During execution, the history at the next chunk $\tilde h_{s_{k+1}}$ is formed using post-blending actions $\tilde{a}_t$ (per Eq.~\eqref{eq:blend}), capturing realistic feedback from prior executions.

To improve robustness and reduce horizon-wise drift, we introduce a temporal continuity prior:
\begin{align}
L_{\text{cont}}
&= \lambda_{\text{TV}} \sum_{t=2}^{T} \|a^{(k)}_t - a^{(k)}_{t-1}\|_1 \\
&\quad + \lambda_{D2} \sum_{t=3}^{T} \|a^{(k)}_t - 2a^{(k)}_{t-1} + a^{(k)}_{t-2}\|_2^2, \nonumber
\end{align}
where the first-order term enforces total variation regularity, and the second-order term penalizes curvature (jerk). This reduces one-step prediction error and suppresses high-frequency instability across chunks. To simulate imperfect rollout scenarios, we apply stochastic corruption to the history during training:
\begin{align}
a^{(k)}_{t-i} &\leftarrow 
\begin{cases}
\bar a_{t-i} + \epsilon,\quad \epsilon \sim \mathcal{N}(0,\sigma^2), & \text{w.p. } 1-q,\\
\mathbf{0}, & \text{w.p. } q,
\end{cases} \notag \\
a^{(k)}_{t-i} &\leftarrow (1{-}\alpha)\,a^{(k)}_{t-i} + \alpha\,\hat a_{t-i},\quad i=1{:}p,
\label{eq:history_dropout}
\end{align}
where $q$ is the dropout rate, and $\alpha\in[0,1]$ is the scheduled sampling ratio. The resulting noisy sequence forms $h_{s_k}$ during training, while clean post-blending actions $\tilde a_t$ are used at test time. We define the supervised training objective as:
\begin{equation}
L_{\text{sup}} = L_{\text{BC}} + L_{\text{cont}} + L_{\text{bdry}},
\end{equation}
where $L_{\text{BC}}$ aligns actions with demonstrations, $L_{\text{cont}}$ enforces local smoothness, and $L_{\text{bdry}}$ (from Sec.~\ref{sec:overlap_blending}) ensures chunk-level consistency. These losses operate on raw predictions $a^{(k)}$, while execution histories are constructed from $\tilde a_t$. Under mild smoothness assumptions, the long-horizon deviation can be bounded as follows. If $\pi_\theta$ is $L_\pi$-Lipschitz and the history noise is bounded by $\epsilon$, with $L_{\text{cont}}$ inducing a contraction $\rho < 1$, the cumulative deviation satisfies:
\begin{equation}
\sum_{t=1}^{T}\|a^{(k)}_t - a^{(k)\star}_t\| \;\le\; \frac{L_\pi}{1 - \rho}\,T\,\epsilon,
\end{equation}
where $a^{(k)\star}_t$ denotes the trajectory under clean histories, establishing the link between local regularization and long-horizon stability.

\subsection{Continuity-Constrained Advantage-Weighted Fine-Tuning}
\label{sec:awac}
While continuity-regularized imitation fosters coherent transitions, it remains limited by the coverage and quality of expert demonstrations. To overcome this, we introduce a reinforcement fine-tuning stage that enables reward-aligned adaptation while preserving the structural priors learned during pretraining. Trajectories are collected using executed (post-blending) actions $\tilde{a}_t$ (Eq.~\eqref{eq:blend}), and training-time histories may be perturbed, as in Eq.~\eqref{eq:history_dropout}, to improve robustness. The policy continues to operate in chunked form, but advantage estimation and updates are performed step-wise within each chunk.

We define the decision state as $s_t = (o_t, \tilde{h}_{s_k}, l)$, where 
$\tilde{h}_{s_k} = (\tilde{a}_{s_k-p}, \dots, \tilde{a}_{s_k-1})$
encodes recently executed actions. We use benchmark-defined sparse success rewards $r_t$ (0/1, terminal for long-horizon tasks). A critic pair $(Q_\phi, V_\phi)$ is trained via temporal-difference regression and expectile value fitting:
\begin{equation}
\label{eq:loss_qv}
\begin{aligned}
L_Q(\phi) &= 
\mathbb{E}\!\left[\left(Q_\phi(s_t, \tilde{a}_t) - (r_t + \gamma V_\phi(s_{t+1}))\right)^2\right], \\
L_V(\phi) &= 
\mathbb{E}\!\left[\rho_{\tau_e}\!\left(A_\phi(s_t, \tilde{a}_t)\right) A_\phi(s_t, \tilde{a}_t)^2 \right], \\
A_\phi(s_t, \tilde{a}_t) &= 
Q_\phi(s_t, \tilde{a}_t) - V_\phi(s_t).
\end{aligned}
\end{equation}
where $\rho_{\tau_e}(u)=|\tau_e - \mathbf{1}\{u < 0\}|$ defines the asymmetric expectile weight. The resulting advantage informs a clipped exponential update weight:
\begin{equation}
w(s_t, \tilde{a}_t) =
\mathrm{clip}\!\left(
\exp\!\left(\max(0, A_\phi(s_t, \tilde{a}_t)) / \tau \right),
1,\; w_{\max}
\right),
\label{eq:adv_weight}
\end{equation}
where $\tau$ is the temperature. Positive-advantage actions are amplified, while suboptimal ones are downweighted. The chunked policy is then updated via:
\begin{equation}
L_{\mathrm{AWAC}}(\theta)
= -\,\mathbb{E}\!\left[
\sum_{t \in \text{chunk}(k)}
w(s_t, \tilde{a}_t)\, \log \pi_\theta(\tilde{a}_t \mid s_t)
\right].
\end{equation}

To maintain temporal consistency, we retain the same regularization used in imitation. Specifically, we apply total variation and curvature penalties, along with a boundary alignment term:
\begin{align}
L_{\mathrm{cont+bdry}}(\theta)
&= \lambda_{\mathrm{TV}}
   \sum_{t \in \text{chunk}(k)}
   \left\|a^{(k)}_t - a^{(k)}_{t-1}\right\|_1  \\
&\quad + \lambda_{D2}
   \sum_{t \in \text{chunk}(k)}
   \left\|a^{(k)}_t - 2a^{(k)}_{t-1} + a^{(k)}_{t-2}\right\|_2^2 \nonumber \\
&\quad + \alpha
   \sum_{t=1}^{O}
   \left\|a^{(k)}_{t} -
          \texttt{sg}\!\left(a^{(k-1)}_{t+L-O}\right)\right\|_2^2. \nonumber
\end{align}
where \texttt{sg} denotes stop-gradient to avoid cyclic dependencies across chunks. These losses operate on predicted actions $a^{(k)}$, while the critic and advantage are evaluated on executed actions $\tilde{a}_t$. The final training objective integrates all components:
\begin{align}
L_{\mathrm{total}}(\theta)
=\;\;& L_{\mathrm{AWAC}}(\theta)
+ L_{\mathrm{cont+bdry}}(\theta)  \notag \\
&+ \beta\,\mathbb{E}_s\!\left[
D_{\mathrm{KL}}\!\big(\pi_\theta(\cdot|s)\,\|\,\pi_{\theta_{\mathrm{old}}}(\cdot|s)\big)
\right]   \notag\\
&- \lambda_H\,\mathbb{E}_s\!\left[\mathcal{H}(\pi_\theta(\cdot|s))\right],
\label{eq:loss_total}
\end{align}
where the KL regularizer and entropy bonus stabilize fine-tuning. This reinforcement stage enables reward-driven improvements without sacrificing temporal smoothness or structural coherence across chunk boundaries.

\section{Experiment}
\label{sec:exp}

\subsection{Dataset and Evaluation Metric}
\label{sec:dataset}

\noindent\textbf{Dataset.}  
We evaluate three instruction-conditioned settings: the standard \textbf{CALVIN} benchmark~\cite{mees2022calvin}, the long-horizon \textbf{LIBERO-Long} benchmark~\cite{zhu2023libero}, and a \textbf{real-world} dataset collected using a 6-DoF collaborative arm~\cite{cytoderm2025}. The real-world setting includes two compositional manipulation tasks: (1) \emph{Cloth Strip Grasping}, which requires sequentially extracting a single strip from a cluttered stack, demanding fine-grained visual discrimination and consistent motion to avoid entanglement; and (2) \emph{Block Insertion}, which involves picking a block from a randomized location and inserting it into a tight fixture, requiring contact-aware control and smooth alignment. Both tasks are instruction-driven and span nontrivial temporal dependencies.
\noindent\textbf{Evaluation Metrics.}
We assess task success and action structure. On {CALVIN}, we report the average episode length (Avg. Len) for 5-step tasks; on {LIBERO-Long}, we report the official long-horizon success rate (Long SR). Action smoothness is measured by first-, second-, and third-order mean squared differences: MSD-$\Delta a = \tfrac{1}{T} \sum_t \|a_t - a_{t-1}\|^2$, MSD-$\Delta^2 a = \tfrac{1}{T} \sum_t \|a_t - 2a_{t-1} + a_{t-2}\|^2$, and MSD-$\Delta^3 a = \tfrac{1}{T} \sum_t \|a_t - 3a_{t-1} + 3a_{t-2} - a_{t-3}\|^2$, capturing velocity, acceleration, and jerk. Seam consistency is evaluated via \textbf{Bjump}, the mean discrepancy in the overlap region $\mathcal{B}$, i.e., $\tfrac{1}{|\mathcal{B}|} \sum_{b \in \mathcal{B}} \|a^{(k)}_b - a^{(k-1)}_b\|$, and \textbf{Bratio}, its normalization by $\tfrac{1}{T} \sum_t \|a_t\|$. Spectral artifacts are measured by high-frequency energy ratio \textbf{HF\_ratio} = $\sum_{\omega \in \Omega_{\text{high}}} |\mathcal{F}(a)(\omega)|^2 / \sum_{\omega} |\mathcal{F}(a)(\omega)|^2$, with $\mathcal{F}(a)$ the Fourier transform. Finally, total variation \textbf{TV-L1} = $\tfrac{1}{T} \sum_t \|a_t - a_{t-1}\|_1$ captures local shifts missed by MSD.

\subsection{Main Results}
\noindent\textbf{Implementation Details.}
Training used four NVIDIA A800s per job (over 32 GPUs across all runs); inference used one A800 in FP16. \textsc{ChunkFlow} adds no inference-time network: deterministic blending is parameter-free and requires no extra policy forward pass, while seam/continuity losses and the AWAC critic are used only during training. Thus, deployment remains a single policy despite the added training cost. To compare controllers with different rollout rates and chunk sizes, we report the \emph{Average Reasoning Latency (ARL)}, the amortized cost per executed action:
$\text{ARL}=\frac{\sum_{c=1}^{K}T_c}{|\mathcal{A}|}$,
where $|\mathcal{A}|=\sum_{c=1}^{K}|\mathbf{a}_c|$. For chunked rollout, $|\mathbf{a}_1|=L$ and $|\mathbf{a}_c|=L-O$ for $c\geq2$.

\noindent\textbf{Comparisons.}
We evaluate \textsc{ChunkFlow} against representative controllers on CALVIN ABC–D, including classic BC/RL, diffusion-based, and autoregressive VLA models. Table~\ref{tab:main_results} reports task success and smoothness metrics across temporal, spectral, and boundary dimensions. Classic methods (e.g., HULC~\cite{nair2023whatmatters}, SPIL~\cite{zhou2024spil}) lack structural modeling, resulting in high-frequency jitter (HF\_ratio $\geq$ 1.0), severe boundary artifacts (Bratio $>$ 2.0), and poor success rates ($\leq$1.7). Diffusion models such as FLOWER~\cite{brohan2024flower} and 3D Diffuser Actor~\cite{zhou2024diffuser} improve global coherence but exhibit large high-order curvature (e.g., MSD-$\Delta^3 a$ $=$ 1.191), reflecting aliasing and seam inconsistency.

VLA baselines (e.g., GR-1~\cite{wu2023gr1}, Seer~\cite{tian2024predictive}) benefit from stronger temporal memory and yield lower second-order variation (MSD-$\Delta^2 a$ $\approx$ 0.16–0.24); yet, they still suffer from unregularized seams, leading to boundary misalignment and unstable long-horizon execution (e.g., Bratio $>$ 2.2). In contrast, \textsc{ChunkFlow} provides a strong smoothness-success trade-off: it achieves competitive task success (4.30), the lowest first-/second-order variation (MSD-$\Delta a$=0.075, MSD-$\Delta^2a$=0.154), Bjump (0.209), HF ratio (0.431), and TV-L1 (0.001), while GR-1 and HULC lead on MSD-$\Delta^3a$ and Bratio. These gains stem from its chunk-structured prediction, overlap-aware blending, and joint regularization of derivatives and seam alignment—enabling smooth, stable, and reward-aligned execution across extended time horizons.

\begin{table*}[t]
\centering
\caption{
\textbf{Main results on CALVIN ABC-D benchmark.} 
We compare task success, multi-level smoothness (temporal,
spectral, boundary), and global variation. Lower is better except Success; bold = column best.
}
\label{tab:main_results}
\resizebox{\linewidth}{!}{
\begin{tabular}{lcccccccc}
\toprule
\textbf{Method} & {Success (Avg. Len.) ↑} & {MSD-$\Delta a$ ↓} & {MSD-$\Delta^2 a$ ↓} & {MSD-$\Delta^3 a$ ↓} & {Bjump ↓} & {Bratio ↓} & {HF\_ratio ↓} & {TV-L1 ↓} \\
\midrule
\midrule
\multicolumn{9}{c}{\emph{Classic BC / RL Baselines (Non-VLM)}} \\
HULC~\cite{nair2023whatmatters}            & 0.67 & 0.117 & 0.296 & 0.956 & 0.241 & \textbf{1.094} & 0.575 & 0.056 \\
SPIL~\cite{zhou2024spil}            & 1.71 & 0.107 & 0.228 & 0.703 & 0.253 & 2.231 & 1.000 & 0.028 \\
\midrule
\multicolumn{9}{c}{\emph{Generative Diffusion / Flow Policies}} \\
3D Diffuser Actor \cite{zhou2024diffuser} & 3.27 & 0.259 & 0.574 & 1.790 & 0.212 & 2.676 & 0.455 & 0.013 \\
VPP \cite{hu2024vpp}      & 4.29 & 0.096 & 0.181 & 0.535 & 0.237 & 1.646 & 1.000 & 0.035 \\
FLOWER \cite{brohan2024flower} & \textbf{4.54} & 0.161 & 0.382 & 1.191 & 0.443 & 2.367 & 0.460 & 0.044 \\
\midrule
\multicolumn{9}{c}{\emph{VLA Policies (Autoregressive / Transformer-based)}} \\
GR-1 \cite{wu2023gr1}      & 3.06 & 0.082 & 0.165 & \textbf{0.496} & 0.969 & 2.224 & 0.999 & 0.006 \\
Seer \cite{tian2024predictive}      & 3.50 & 0.091 & 0.186 & 0.556 & 0.297 & 1.957 & 0.442 & 0.007 \\
UniVLA \cite{belkhale2024minivla} & 3.80 & 0.118 & 0.238 & 0.714 & 0.251 & 4.109 & 0.488 & 0.009 \\
\midrule
\textbf{ChunkFlow (\textcolor{red}{Ours})}  & {4.30} & \textbf{0.075} & \textbf{0.154} & {0.512} & \textbf{0.209} & {1.471} & \textbf{0.431} & \textbf{0.001} \\
\bottomrule
\end{tabular}}
\vspace{-0.2cm}
\end{table*}

\noindent\textbf{Cross-Dataset Evaluation on Long-Horizon Tasks.}  
We evaluate generalization on LIBERO~\cite{zhu2023libero}, which emphasizes long-horizon task compositionality and layout shifts. As shown in Table~\ref{tab:cross_dataset}, \textsc{ChunkFlow} achieves \textbf{93.4\%} success—slightly outperforming PI0.5~\cite{black2025pi05} (92.6\%)—and outperforms OpenVLA (53.7\%), VPP (38.9\%), and CLIP-RT (83.8\%). While PI0.5-RTC~\cite{black2025real} applies inference-time chunk blending and reduces boundary artifacts (Bjump: 0.167$\rightarrow$0.115, HF\_ratio: 0.494$\rightarrow$0.342), its success drops sharply to \textbf{83.7\%}. This supports our hypothesis: inference-only smoothing cannot resolve misaligned seam predictions, causing bias accumulation and degraded rollout stability.

By contrast, \textsc{ChunkFlow} enforces continuity during training and directly optimizes boundary alignment across paired chunks. It achieves the lowest motion deviation (MSD-$\Delta a$ = \textbf{0.042}, $\Delta^2 a$ = \textbf{0.197}, $\Delta^3 a$ = \textbf{0.235}), minimal discontinuities (Bjump = \textbf{0.082}), and the cleanest temporal spectrum (HF\_ratio = \textbf{0.135}, TV-L1 = \textbf{0.011}). Its average action inference latency is also the lowest (\textbf{4.43 ms}) due to amortized inference over fixed-step overlapping chunks ($S=L{-}O$), which reduces sampling redundancy and stabilizes per-action rollout. In contrast, RTC incurs extra planning costs from heuristic smoothing and late-stage proposal reweighting, resulting in a higher ARL (18.47\,ms). These results demonstrate that seam-aware \emph{training}, not inference heuristics, is critical for ensuring temporal smoothness and efficient execution under long-horizon tasks.

\begin{table*}[t]
\centering
\caption{
\textbf{Cross-Dataset Generalization on LIBERO benchmark.}
We evaluate long-horizon manipulation on LIBERO to assess generalization beyond CALVIN. 
\textsc{ChunkFlow} achieves strong success while suppressing discontinuities and high-frequency jitter. We also report \textbf{ARL (ms)}, the average action inference latency. Lower is better for all smoothness and latency metrics.
}
\label{tab:cross_dataset}
\resizebox{\linewidth}{!}{
\begin{tabular}{lccccccccc}
\toprule
\textbf{Method} & Long SR (\%) $\uparrow$ & MSD-$\Delta a$ ↓ & MSD-$\Delta^2 a$ ↓ & MSD-$\Delta^3 a$ ↓ & Bjump ↓ & Bratio ↓ & HF\_ratio ↓ & TV-L1 ↓ & ARL (ms) ↓ \\
\midrule
\midrule
OpenVLA \cite{kim2024openvla}     & 53.7 & 0.083 & 0.229 & 0.743 & 0.166 & 1.145 & 0.862 & 0.031 & 219.43 \\
VPP \cite{hu2024vpp}              & 38.9 & 0.133 & 0.255 & 0.732 & 0.738 & 5.702 & 0.394 & 0.029 & 13.91 \\
PI0.5 \cite{black2025pi05}       & {92.6} & 0.095 & 0.249 & 0.812 & 0.167 & 1.401 & 0.494 & 0.023 & 9.04 \\
PI0.5-RTC \cite{black2025real}    & 83.7 & 0.089 & 0.240 & 0.729 & 0.115 & 0.850 & 0.342 & 0.018 & 18.47 \\
CLIP-RT \cite{kang2024clip}       & 83.8 & 0.135 & 0.302 & 0.916 & 0.179 & 1.350 & 0.999 & 0.026 & 6.86 \\
Seer \cite{tian2024predictive}    & 87.7 & 0.141 & 0.279 & 0.639 & 0.195 & 1.602 & 0.337 & 0.022 & 14.25 \\
\midrule
\textbf{ChunkFlow (\textcolor{red}{Ours})} & \textbf{93.4} & \textbf{0.042} & \textbf{0.197} & \textbf{0.235} & \textbf{0.082} & \textbf{0.082} & \textbf{0.135} & \textbf{0.011} & \textbf{4.43} \\
\bottomrule
\end{tabular}}
\vspace{-0.4cm}
\end{table*}

\subsection{Ablation Study}

\begin{figure}[t]
    \centering
    \includegraphics[width=\linewidth]{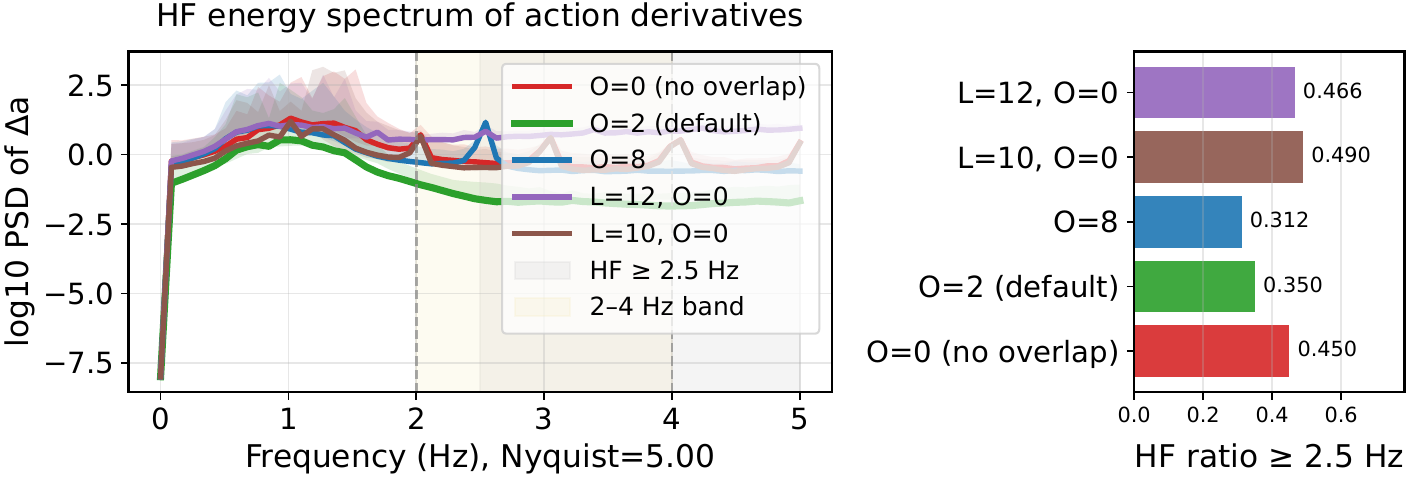}
    \vspace{-0.2cm}
    \caption{
    \textbf{High-frequency (HF) energy spectrum under varying overlap and chunk length.}
    We plot the log-scaled power spectral density (PSD) of action derivatives to assess smoothness. The high-frequency band (2.5–5.0\,Hz) reflects unstable transitions. The default \((L{=}10, O{=}8)\) maintains low HF energy beyond 3\,Hz, while no overlap \((O{=}0)\) and long chunks \((L{=}12)\) induce sharp spikes and spectral bursts. Shaded areas denote standard error across episodes and dimensions.
    }
    \label{fig:hf_plot}
    \vspace{-0.2cm}
\end{figure}

\noindent\textbf{Hyperparameter Sensitivity}.
We validate our theoretical predictions on continuity-consistent execution by probing architectural and regularization factors. Seam errors grow with stride \((L{-}O)\), and increasing overlap improves pre-blending agreement and suppresses aliasing: \(O{=}0\) yields high-frequency artifacts (HF\_ratio = 0.500; Bjump = 0.230), while moderate \(O{=}4\) minimizes the spectrum (HF\_ratio = 0.371), and executable \(O{=}8\) balances spectral stability (MSD-$\Delta^2 a$ = 0.154) with the best seam coherence (Bjump = 0.209). Chunk horizon also matches expectations: too-short \(L{=}8\) removes temporal context, harming acceleration smoothness (MSD-$\Delta^2 a$ = 0.205), while too-long \(L{=}12\) amplifies extrapolation error, producing bursts (HF\_ratio = 0.812). Fig.~\ref{fig:hf_plot} confirms that only \((L{=}10,O{=}8)\) suppresses energy beyond 3\,Hz. Regularizers exhibit predictable failure modes: over-strong curvature (\(\lambda_{D2}{=}0.007\)) induces phase lag and seam drift (MSD-$\Delta a$ = 0.121; Bjump = 0.214), excessive boundary loss increases curvature (MSD-$\Delta^2 a$ = 0.209; Bjump = 0.221), and a large sparsity prior (\(3{\times}10^{-4}\)) harms smoothness without spectral gain. Raising TV (\(\lambda_{\text{TV}}\!:\!0.005\!\to\!0.015\)) suppresses natural variation, degrading both $\Delta a$ (0.075→0.091) and $\Delta^2 a$ (0.154→0.170). These trends support our core insight: moderate overlap and chunk length reduce stride-induced aliasing, while over-regularization destabilizes dynamics through curvature amplification and delayed transitions.

\begin{table}[t]
\centering
\caption{
\textbf{One-Factor Hyperparameter Sensitivity.}
We vary each factor individually while fixing others to assess robustness. Moderate values yield the best balance between task success, smoothness, and seam consistency. Lower is better except for Success.
}
\label{tab:ablation_hyper}
\resizebox{\linewidth}{!}{
\begin{tabular}{lccccc}
\toprule
\textbf{Setting (1-factor)} & Success $\uparrow$ & MSD-$\Delta a$ $\downarrow$ & MSD-$\Delta^2 a$ $\downarrow$ & Bjump $\downarrow$ & HF\_ratio $\downarrow$ \\
\midrule
\midrule
\multicolumn{6}{l}{\emph{First-order regularizer \(\lambda_{\text{TV}}\) (fix \(L,O,\lambda_{D2},\lambda_{B},\lambda_{\text{prior}}\))}} \\
\quad \(\lambda_{\text{TV}}{=}0.005\) (default)    & \textbf{4.30} & \textbf{0.075} & \textbf{0.154} & \textbf{0.209} & \textbf{0.431} \\
\quad \(\lambda_{\text{TV}}{=}0.008\)              & 4.29 & 0.083 & 0.162 & 0.214 & 0.433 \\
\quad \(\lambda_{\text{TV}}{=}0.015\)              & 4.26 & 0.091 & 0.170 & 0.218 & 0.437 \\
\midrule
\multicolumn{6}{l}{\emph{Second-order regularizer \(\lambda_{D2}\) (fix \(L,O,\lambda_B,\lambda_{\text{prior}}\))}} \\
\quad \(\lambda_{D2}{=}0.003\)        & 4.26 & 0.100 & 0.170 & 0.210 & 0.455 \\
\quad \(\lambda_{D2}{=}0.005\) (default) & \textbf{4.30} & \textbf{0.075} & \textbf{0.154} & \textbf{0.209} & \textbf{0.431} \\
\quad \(\lambda_{D2}{=}0.007\)        & 4.25 & 0.121 & 0.206 & 0.214 & 0.472 \\
\midrule
\multicolumn{6}{l}{\emph{Boundary loss \(\lambda_{B}\) (fix \(L,O,\lambda_{D2},\lambda_{\text{prior}}\))}} \\
\quad \(\lambda_{B}{=}0.02\)          & 4.27 & 0.096 & 0.165 & 0.222 & 0.437 \\
\quad \(\lambda_{B}{=}0.03\) (default)& \textbf{4.30} & \textbf{0.075} & \textbf{0.154} & \textbf{0.209} & \textbf{0.431} \\
\quad \(\lambda_{B}{=}0.04\)          & 4.23 & 0.122 & 0.209 & 0.221 & 0.432 \\
\midrule
\multicolumn{6}{l}{\emph{Sparsity prior \(\lambda_{\text{prior}}\) (fix \(L,O,\lambda_{D2},\lambda_{B}\))}} \\
\quad \(1{\times}10^{-4}\)            & 4.26 & 0.097 & 0.168 & 0.217 & 0.437 \\
\quad \(2{\times}10^{-4}\) (default)  & \textbf{4.30} & \textbf{0.075} & \textbf{0.154} & \textbf{0.209} & \textbf{0.431} \\
\quad \(3{\times}10^{-4}\)            & 4.24 & 0.117 & 0.202 & 0.213 & 0.433 \\
\midrule
\multicolumn{6}{l}{\emph{Overlap ratio \(O/L\) (retrain; fix \(L,\lambda_{D2},\lambda_{B},\lambda_{\text{prior}}\))}} \\
\quad \(O{=}0\)                       & 4.20 & 0.118 & 0.200 & 0.230 & 0.500 \\
\quad \(O{=}2\)                       & 4.10 & 0.105 & 0.175 & 0.450 & {0.470} \\
\quad \(O{=}4\)                       & 4.20 & 0.108 & 0.179 & 0.388 & 0.371 \\
\quad \(O{=}8\) (default)             & \textbf{4.30} & \textbf{0.075} & \textbf{0.154} & \textbf{0.209} & \textbf{0.431} \\
\midrule
\multicolumn{6}{l}{\emph{Chunk length \(L\) (retrain; fix \(O,\lambda_{D2},\lambda_{B},\lambda_{\text{prior}}\))}} \\
\quad \(L{=}8\)                       & 4.30 & 0.120 & 0.205 & 0.205 & 0.450 \\
\quad \(L{=}10\) (default)            & \textbf{4.30} & \textbf{0.075} & \textbf{0.154} & \textbf{0.209} & \textbf{0.431} \\
\quad \(L{=}12\)                      & 4.20 & 0.102 & 0.192 & 0.240 & 0.812 \\
\bottomrule
\end{tabular}}
\vspace{-0.2cm}
\end{table}

\noindent\textbf{Inference-Time Stability and Smoothness Enforcement.}  
To assess deployment robustness, we examine test-time strategies for enforcing temporal consistency in chunked rollouts. We first evaluate naive overlap execution without retraining (Table~\ref{tab:abl_infer_stab}, A1): small overlap ($O{=}2$) slightly reduces curvature (MSD-$\Delta^2 a$: 0.381 $\to$ 0.360), but larger values ($O{=}4/8$) degrade both smoothness and success due to redundant predictions and conflicts across chunks, confirming that inference-only overlap suffers from gradient ambiguity under perceptual drift. We next study lightweight smoothing heuristics (A2): blending ($O{=}2$), EMA filtering, and warm-started chunk latents. Each improves specific metrics (e.g., warm-start lowers MSD-$\Delta^2 a$ to 0.163), but their combination yields the best stability (HF\_ratio = 0.431, Bjump = 0.209, Success = 4.30), even when trained with $O{=}0$. We further simulate latency via $\pm$100\,ms control delay (B), observing negligible degradation—indicating strong temporal generalization. Finally, classical filters (Table~\ref{tab:ablation_filter_main}) like Savitzky–Golay and Butterworth reduce high-order jitter (e.g., MSD-$\Delta^3 a = 0.002$); however, they fail to suppress seam artifacts (Bjump $\geq$ 0.228) and spectral aliasing (HF\_ratio $\geq$ 0.69). These results highlight that post-hoc smoothing cannot resolve structural discontinuities. In contrast, ChunkFlow enforces global coherence via derivative-aligned chunk execution learned through training-time regularization.

\begin{table}[t]
\centering
\caption{\textbf{Inference-time stability with overlap smoothing and control jitter.}  
We ablate naive overlap blending (A1), smoothing strategies (A2), and control delay robustness (B). Lower is better except Success.}
\label{tab:abl_infer_stab}
\resizebox{\linewidth}{!}{
\begin{tabular}{lcccc}
\toprule
\textbf{Setting} & Success $\uparrow$ & MSD-$\Delta^2 a$ $\downarrow$ & HF\_ratio $\downarrow$ & Bjump $\downarrow$ \\
\midrule
\midrule
\multicolumn{5}{l}{\emph{A1. Naive Chunk Overlap (no retraining)}} \\
Overlap ($O{=}0$)               & 4.29 & 0.381 & 1.000 & 0.437 \\
Overlap ($O{=}2$)               & 4.10 & 0.360 & 0.950 & 0.410 \\
Overlap ($O{=}4$)               & 3.95 & 0.390 & 0.940 & 0.385 \\
Overlap ($O{=}8$)               & 3.60 & 0.440 & 0.970 & 0.420 \\
\midrule
\multicolumn{5}{l}{\emph{A2. Lightweight Smoothing (no retraining)}} \\
No smoothing                   & 4.29 & 0.273 & 0.445 & 0.493 \\
\quad + Blend ($O{=}2$)        & 4.21 & 0.317 & 0.501 & 0.299 \\
\quad + EMA ($\alpha{=}0.1$)  & 4.27 & 0.273 & 0.441 & 0.453 \\
\quad + EMA ($\alpha{=}0.3$)  & 4.27 & 0.283 & 0.438 & 0.376 \\
\quad + Warm start            & 4.31 & 0.163 & 0.329 & 0.428 \\
\rowcolor{gray!08}
Blend + EMA ($\alpha{=}0.3$) + Warm start & \textbf{4.30} & \textbf{0.154} & \textbf{0.431} & \textbf{0.209} \\
\midrule
\multicolumn{5}{l}{\emph{B. Delay Robustness (jitter $\pm j$\,ms)}} \\
$\pm$0 ms                      & 4.31 & 0.222 & 0.392 & 0.202 \\
$\pm$10 ms                     & 4.31 & 0.220 & 0.392 & 0.199 \\
$\pm$50 ms                     & 4.30 & 0.222 & 0.390 & 0.203 \\
\rowcolor{gray!08}
$\pm$100 ms                    & \textbf{4.30} & \textbf{0.221} & \textbf{0.391} & \textbf{0.203} \\
\bottomrule
\end{tabular}}
\end{table}

\begin{table}[t]
\centering
\caption{
\textbf{Comparison of post-hoc filters versus training-time smoothness integration.}
While post-processing (e.g., Savitzky–Golay, Butterworth) improves local smoothness, it fails to suppress high-frequency noise and seam discontinuities.
ChunkFlow enforces smoothness during training, achieving superior temporal and spectral coherence.
Lower is better.
}
\label{tab:ablation_filter_main}
\resizebox{\linewidth}{!}{
\begin{tabular}{lccccl}
\toprule
\textbf{Filter} & MSD-$\Delta^3 a$ $\downarrow$ & HF\_ratio $\downarrow$ & Bjump $\downarrow$ & Notes \\
\midrule
\midrule
\textcolor{gray}{Base (raw)}         & \textcolor{gray}{0.839} & \textcolor{gray}{1.000} & \textcolor{gray}{0.493} & \textcolor{gray}{No smoothing} \\
Savitzky–Golay        & 0.045 & 0.78 & 0.256 & Post-hoc \\
Butterworth (low-pass)& 0.002 & 0.69 & 0.228 & Post-hoc \\
Cubic spline          & 0.839 & 0.97 & 0.493 & Offline fitting \\
\rowcolor{gray!10}
\textbf{ChunkFlow (ours)} & \textbf{0.512} & \textbf{0.431} & \textbf{0.209} & Integrated during training \\
\bottomrule
\end{tabular}}
\end{table}

\noindent\textbf{History Length Sensitivity.}  
We evaluate how varying temporal context ($p$) impacts prediction stability. Without history ($p{=}0$), the policy struggles with acceleration smoothness (MSD-$\Delta^2 a{=}0.223$), third-order coherence (MSD-$\Delta^3 a{=}0.582$), and boundary stability (Bjump$=0.241$), due to reliance on instantaneous observations. Short histories ($p{=}2$) yield clear improvements (MSD-$\Delta^2 a{=}0.207$, Bjump$=0.220$), while moderate history ($p{=}4$) achieves optimal results across all axes (e.g., HF\_ratio$=0.431$, success$=4.30$), enabling consistent chunk transitions. However, longer context ($p{=}8$) degrades performance (MSD-$\Delta^2 a{=}0.203$, Bjump$=0.224$) as outdated signals introduce error accumulation. These trends confirm that a limited, recent history improves alignment and smoothness, but excessive memory leads to temporal drift.

\begin{table}[t]
\centering
\caption{\textbf{Effect of history length $p$.} Increasing temporal context initially improves smoothness and boundary stability, but excessive history introduces error accumulation. Lower is better except Success.}
\label{tab:ablation_historylen}
\resizebox{\linewidth}{!}{
\begin{tabular}{lccccc}
\toprule
$p$ & Success $\uparrow$ & MSD-$\Delta^2 a$ $\downarrow$ & MSD-$\Delta^3 a$ $\downarrow$ & HF\_ratio $\downarrow$ & Bjump $\downarrow$ \\
\midrule
\midrule
0 & 4.29  & 0.223  & 0.582  & 0.462  & 0.241 \\
2 & 4.27  & 0.207  & 0.545  & 0.431  & 0.220 \\
\rowcolor{gray!08}\textbf{4} & \textbf{4.30} & \textbf{0.154} & \textbf{0.512}  & \textbf{0.431} & \textbf{0.209} \\
8 & 4.13  & 0.203  & 0.538  & 0.450  & 0.224 \\
\bottomrule
\end{tabular}}
\end{table}

\noindent\textbf{Policy Adaptation with Safety Constraints.}  
To enhance long-horizon performance beyond supervised pretraining, we apply reinforcement fine-tuning using AWAC, with gradient clipping and imitation anchoring. As shown in Table~\ref{tab:ablation_rl_merged}, the supervised-only model produces unstable execution with strong high-frequency jitter (HF\_ratio = 1.000), sharp curvature (MSD-$\Delta^3 a$ = 0.535), and seam misalignment (Bjump = 0.237). Adding unconstrained RL (AWAC w/o continuity) reduces spectral noise (↓28\%) but fails to correct boundary inconsistencies. In contrast, our Safe RL-FT achieves coherent adaptation: MSD-$\Delta^3 a$ drops to 0.512, HF\_ratio to 0.431, and Bjump to 0.209, recovering full task success (4.30). These results confirm that continuity-aware regularization is critical for reward-driven updates to avoid seam drift and maintain rollout smoothness.

\begin{table}[t]
\centering
\caption{\textbf{Safe RL fine-tuning.} We ablate reinforcement learning strategies under different update constraints. ChunkFlow benefits from clipped gradients and BC regularization, achieving smoother and more stable rollouts. Lower is better except Success.}
\label{tab:ablation_rl_merged}
\resizebox{\linewidth}{!}{
\begin{tabular}{lcccc}
\toprule
\textbf{Method / Variant} & {Success $\uparrow$} & {MSD-$\Delta^3 a$ $\downarrow$} & {HF\_ratio $\downarrow$} & {Bjump $\downarrow$} \\
\midrule
\midrule
Supervised only (no RL)                & 4.10 & 0.535 & 1.000 & 0.237 \\
AWAC (w/o continuity)                  & 4.17 & 0.530 & 0.720 & 0.225 \\
\rowcolor{gray!08}\textbf{Safe RL-FT (default)} & \textbf{4.30} & \textbf{0.512} & \textbf{0.431} & \textbf{0.209} \\
\bottomrule
\end{tabular}}
\end{table}

\begin{figure}[t]
\centering
\includegraphics[width=\linewidth]{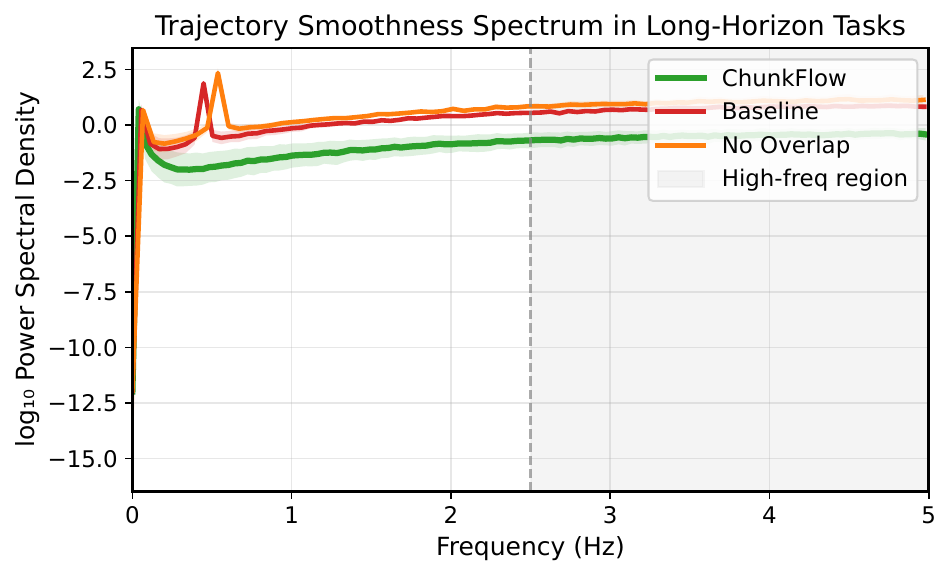}
\caption{
\textbf{Trajectory smoothness spectrum.}
\textsc{ChunkFlow} suppresses high-frequency spectral energy ($>$2.5\,Hz), indicating stable action derivatives in long-horizon tasks. Baselines without structural regularization exhibit aliasing and jitter, visible as sharp PSD peaks.
}
\label{fig:smoothness_spectrum}
\vspace{-0.2cm}
\end{figure}

\begin{figure}[t]
  \centering
  \includegraphics[width=1.05\linewidth]{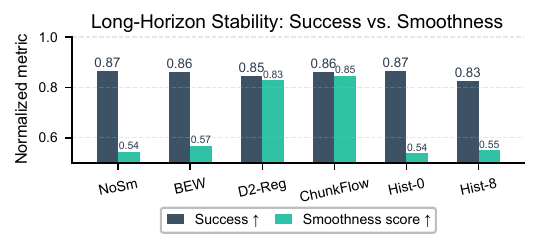}
  \caption{
    \textbf{Trade-off between trajectory smoothness and task performance.}
    Bars report normalized Success and Smoothness ($1{-}\mathrm{HF\_ratio}$) across key variants:
    NoSm (no smoothing), BEW (Blend+EMA+Warm), D2-Reg (with D2 loss), ChunkFlow (ours), Hist-0, and Hist-8.
    \textsc{ChunkFlow} lies on the Pareto frontier, balancing high success with smooth execution.
  }
  \label{fig:pareto}
  \vspace{-0.2cm}
\end{figure}

\noindent\textbf{Long-Horizon Stability.}  
We assess \textsc{ChunkFlow}'s robustness in long-horizon rollouts, where autoregressive policies often accumulate aliasing and drift. Fig.~\ref{fig:smoothness_spectrum} shows the power spectral density (PSD) of action derivatives, highlighting that \textsc{ChunkFlow} significantly suppresses high-frequency components beyond 2.5\,Hz—unlike baselines lacking seam-aware blending or curvature regularization, which exhibit spectral spikes due to discontinuities. This confirms that our structural constraints effectively mitigate jitter over extended horizons. To evaluate the trade-off between smoothness and effectiveness, Fig.~\ref{fig:pareto} plots normalized task Success versus Smoothness ($1{-}\mathrm{HF\_ratio}$). \textsc{ChunkFlow} achieves the best balance (0.86, 0.57), outperforming post-hoc filters like BEW (0.82, 0.66) and long-history variants such as Hist-8 (0.83, 0.45). These results show that long-horizon stability cannot be achieved via denoising alone; structural continuity is essential. \textsc{ChunkFlow}'s overlap blending and trajectory-aware training yield executions that remain stable \textbf{over time}.

\section{CONCLUSIONS}
We introduced {ChunkFlow}, a continuity-consistent framework for chunked robotic policy execution. Rather than stacking isolated techniques, {ChunkFlow} realizes a unified execution model in which overlap-aware blending, seam-regularized learning, and advantage-weighted adaptation are jointly optimized to correct boundary discontinuities in chunked VLA and generative controllers. Experiments on CALVIN, LIBERO, and real-robot benchmarks show improved smoothness, temporal stability, and long-horizon success while preserving the low latency of chunked inference. These results support execution-indexed chunk alignment in simulation and initial hardware tests, while broader real-world robustness remains future work.





\section*{ACKNOWLEDGMENT}
This work was supported by the Natural Science Foundation of China under Grant NSFC 62573343 and 62088102, National Basic Strengthen Research Program of ReRAM under Grant2022-00-03, Fundamental Research Funds for the Central Universities xzy012024066,and by the Open-End Fund of Beijing Institute of Control Engineering under GrantOBCandETL-2024-04.


\bibliographystyle{unsrt}
\bibliography{references}

\end{document}